\documentclass{article}

     \PassOptionsToPackage{numbers, compress}{natbib}

    \usepackage[preprint]{neurips_2019}

\usepackage[utf8]{inputenc} 
\usepackage[T1]{fontenc}    
\usepackage{hyperref}       
\usepackage{url}            
\usepackage{booktabs}       
\usepackage{amsfonts}       
\usepackage{nicefrac}       
\usepackage{microtype}      
\usepackage{listings}       
\usepackage{graphicx}       
\usepackage{float}
\usepackage{subcaption}
\usepackage{color}          
\usepackage{lipsum}

\definecolor{deepblue}{rgb}{0,0,0.5}
\definecolor{deepred}{rgb}{0.6,0,0}
\definecolor{deepgreen}{rgb}{0,0.5,0}
\definecolor{gray}{rgb}{0.3,0.3,0.3}

\title{Geomancer: An Open-Source Framework for Geospatial Feature Engineering}

\author{%
  \textbf{Lester James V. Miranda, Mark Steve Samson, Alfiero K. Orden II}\\
  \textbf{Bianca S. Silmaro, Ram K. De Guzman III, Stephanie S. Sy}\\
  Thinking Machines Data Science\\
  Metro Manila, Philippines\\
  \texttt{\{lj,marksteve,ardie,bianca,ram,stef\}@thinkingmachin.es} \\
}

\begin{document}

\maketitle

\begin{abstract}
    This paper presents Geomancer, an open-source framework for geospatial
    feature engineering. It simplifies the acquisition of geospatial attributes
    for downstream, large-scale machine learning tasks.  Geomancer leverages
    any geospatial dataset stored in a data warehouse\textemdash users need
    only to define the features (\textit{Spells}) they want to create, and cast
    them on any spatial dataset. In addition, these features can be exported
    into a JSON file (\textit{SpellBook}) for sharing and reproducibility.
    Geomancer has been useful to some of our production use-cases such as
    property value estimation, area valuation, and more. It is available on
    Github\footnote{\url{https://github.com/thinkingmachines/geomancer}}, and
    can be installed from PyPI.
\end{abstract}

\section{Introduction}

Geospatial data allows us to gain massive amounts of information based on
location. When samples\textemdash expressed as points, polygons, or rasters
with real-world coordinates\textemdash are coupled with large-scale datasets
such as OpenStreetMap (OSM) \cite{osm2017}, we can gain an information-rich
dataset to derive insights from. To illustrate, given your current position, it
is possible to obtain, say, the number of malls within 1.5-km, the distance to
the nearest supermarket, or the frequency of traffic jams\textemdash all of
which can be used later on for downstream machine learning tasks. 

However, engineering features for geospatial data is a challenging task,
requiring significant amount of compute and storage
\cite{nargesian2017learning, nargesian2018dataset, storcheus2015survey}.
Important considerations include (1) the storage capacity to house geospatial
data sources, (2) the compute complexity to query from that source, and the (3)
ease of extracting information from these sources \cite{klien2005requirements}. 

In this paper, we introduce Geomancer, an open-source framework to perform
geospatial feature engineering at scale. It leverages a data warehouse,
geospatial datasets, and a Python library to pull out information from spatial
datasets. In addition, Geomancer provides a solution for versioning and sharing
feature transforms for other users. It is open-source and licensed under MIT.
Geomancer has been used for production machine learning use-cases such as area
valuation, poverty mapping, and real-estate price estimation.

\section{Architecture}

\paragraph{Concepts}

The fundamental unit in Geomancer is a logical feature \cite{smith2017ballet}
called a \textit{Spell}. It maps a coordinate into a vector of feature values,
$f_{j}^{\mathcal{D}} : \mathcal{V}^{2} \rightarrow \mathbb{R}^{q_j}$, where
$\mathcal{V}$ is the set of feasible coordinates (latitude and longitude in
EPSG:4326 \cite{WGS84EPS46:online}) and $q_{j}$ is the dimensionality of the
$j$th feature vector. A collection of spells, i.e., a \textit{SpellBook}, is
then defined as a set of feature functions $\mathcal{F}^{\mathcal{D}} = \{f_j
\vert j = 1 \dots m\}$. 

\begin{figure}[t]
    \begin{center}
        \includegraphics[width=0.85\linewidth]{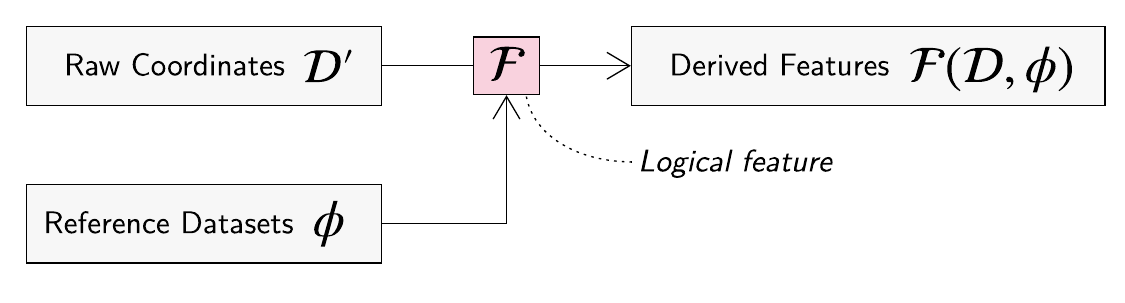}
    \end{center}
    \caption{
        \textit{Information flow in the Geomancer framework.}  In order to obtain
        the features needed for modelling, a set of raw coordinates and
        reference datasets (e.g. OpenStreetMap, Waze, etc.) are supplied to the
        logical feature functions.
    }
  \label{fig:info_flow}
\end{figure}

Figure \ref{fig:info_flow} shows the information flow in the Geomancer
framework. Given a reference data source $\phi$, Geomancer allows users to
define feature transforms $F^{\mathcal{D}}$, and apply these functions to a
dataset containing spatial coordinates $\mathcal{D}^{\prime}$. The result is a
feature matrix $X^{\mathcal{D}^\prime}$ \cite{smith2017ballet} that can be used
for downstream machine learning tasks: 

\begin{equation}
    X^{\mathcal{D}^\prime} = 
    \mathcal{F}^{\mathcal{D}}(\mathcal{D}^\prime, \phi) = 
    (f_{1}^{\mathcal{D}}(\mathcal{D}^\prime, \phi), \dots,
    f_{m}^{\mathcal{D}}(\mathcal{D}^\prime, \phi))
\end{equation}

\begin{figure}[t]
    \begin{center}
        \includegraphics[width=0.85\linewidth]{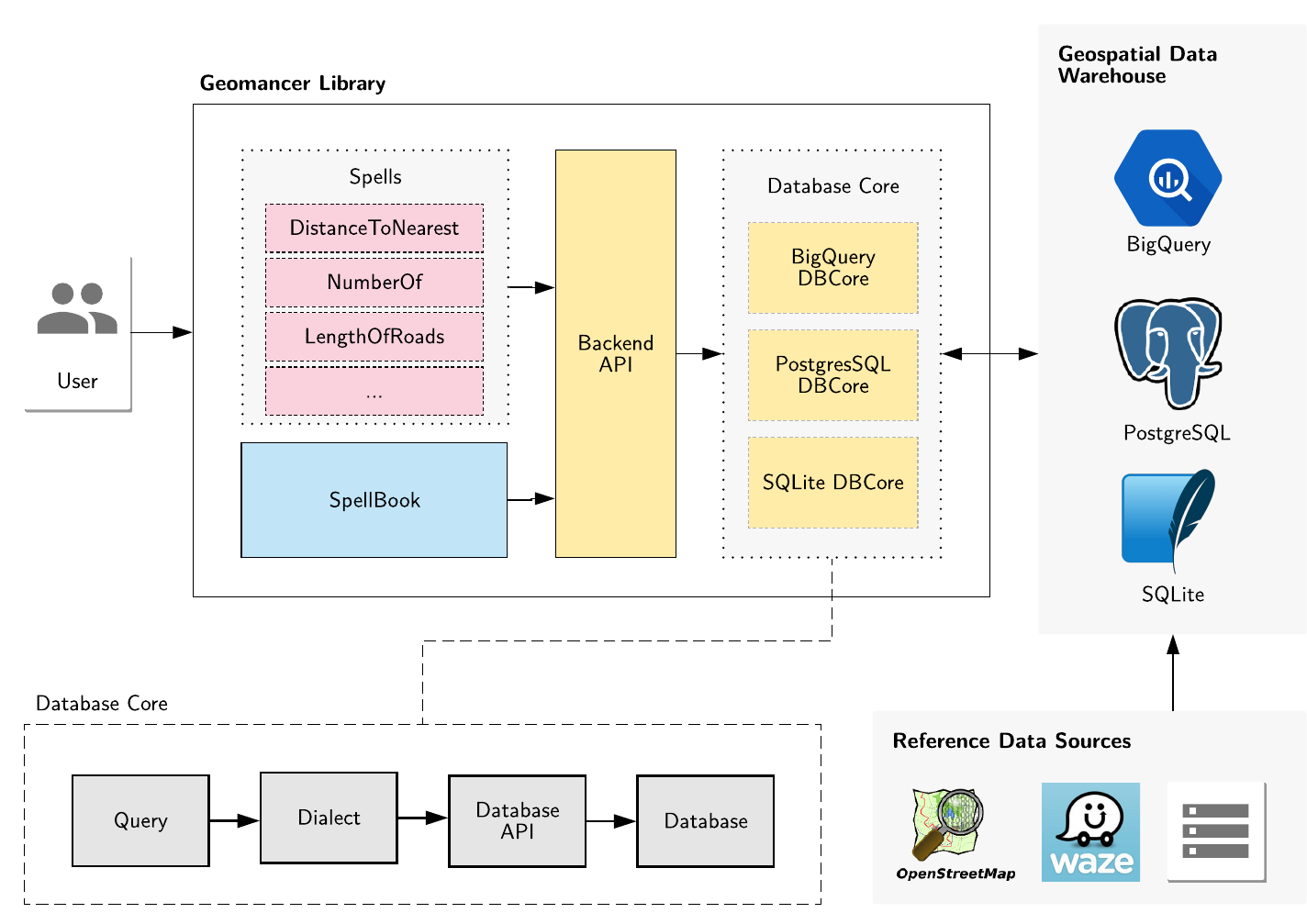}
    \end{center}
    \caption{
      \textit{Architecture of the Geomancer framework.} The user defines
      feature functions $\mathcal{F}$  (Spells or SpellBook) via a Python
      library, which are then executed by the \texttt{DBCore} as query dialects
      at runtime. These function-calls then interact with a data warehouse to
      fetch and store the extracted features. As of now, Geomancer supports
      BigQuery, PostgreSQL, and SQLite for storage with OpenStreetMap (OSM) and
      Waze as a reference data source. 
    }
  \label{fig:architecture}
\end{figure}

\paragraph{System Design} There are three main components in the Geomancer
framework: a Python library client, a data warehouse server, and a reference
data source (Figure \ref{fig:architecture}). 

\begin{itemize}
    \item \textit{Python library client} The \texttt{geomancer} library
        \footnote{\url{https://pypi.org/projects/geomancer}} serves as the
        framework's user-interface. Users can define feature functions
        (Spells or SpellBooks), export/read SpellBooks, and apply transforms to
        any given spatial dataset. Creating new features is done via the
        factory design pattern whereas the SpellBook mechanism is accomplished
        using the builder pattern \cite{gamma1995design}.
    \item \textit{Data warehouse server} The data warehouse provides the
        storage and compute capacity in the framework. The library client can
        connect to multiple databases at the same time, and can handle both
        online transactional (OLTP) or analytical (OLAP) processing workloads.
        The extracted features can be stored inside the warehouse or exported
        as a dataframe for immediate consumption.
    \item \textit{Reference data source} A reference data source is loaded
        inside the warehouse as basis for feature engineering. For example, if
        we want to obtain the number of malls within a 1.5-km radius, we should
        have some knowledge of all mall locations within the area in question.
        Fortunately, open datasets such as OpenStreetMap (OSM) \cite{osm2017}
        exists to give such information. Usually, we create
        Extract-Transform-Load (ETL) pipelines to deliver timely, rich, and
        accurate data from external sources. 
\end{itemize}

\section{Usage}

In practice, Geomancer enables researchers to (1) define geospatial features
for extraction, (2) connect to various data warehouses, and (3) replicate and
version features.  The following sections will demonstrate how this can be done
in the framework.

\paragraph{Feature functions for geospatial feature engineering}
 A \texttt{Spell} provides a declarative interface to define logical features
 \cite{smith2017ballet}. They can be \texttt{cast}ed to a set of coordinates
 after instantiation.  For example, if we wish to get the distance to the
 nearest embassy given a sample of coordinates, we write the following:

\begin{lstlisting}[language=Python]
from geomancer.spells import DistanceToNearest
from tests.conftest import sample_points

# Load a sample of points as a DataFrame
df = sample_points()

# Define a spell
spell = DistanceToNearest("embassy",
                          source_table="ph_osm.gis_osm_pois_free_1",
                          dburl="bigquery://geospatial",
                          feature_name="dist_embassy")

# Cast the spell
df_with_features = spell.cast(df)
\end{lstlisting}

\paragraph{Connect to various data warehouses}
Geomancer can establish a connection to any warehouse by providing a valid
database URL. In practice, this feature has been helpful when engineering
features across tables from different locations (e.g., OSM dataset is stored in
BigQuery, traffic dataset in PostGIS, etc.). So far, Geomancer supports the
following database backends:

\begin{itemize}
    \item BigQuery, an analytics data warehouse from the Google Cloud
        Platform \cite{google2012bigquery, melnik2010dremel}.
    \item PostGIS, a geospatial extension for PostgreSQL
        \cite{stonebraker1987postgres, stonebraker1986design}.
    \item SpatiaLite, a geospatial extension for SQLite \cite{bhosale2015sqlite,
        spatialite}.
\end{itemize}

\paragraph{Save and share feature functions}
Features can be grouped together to form a \texttt{SpellBook}, allowing us to
cast multiple \texttt{Spells} at once. In addition, \texttt{SpellBooks} can be
exported into a JSON file with various metadata (e.g., author, description,
etc.) regarding the feature collection:

\begin{lstlisting}[language=Python]
from geomancer.spells import DistanceToNearest, NumberOf
from geomancer.spellbook import SpellBook

# Create a spellbook
my_spellbook = SpellBook(
          spells=[
              DistanceToNearest("primary",
                                 dburl="bigquery://geospatial",
                                 source_table="ph_osm.gis_osm_roads_free_1",
                                 feature_name="dist_primary"),
              NumberOf("supermarket"
                        dburl="bigquery://geospatial",
                        source_table="geospatial.ph_osm.gis_osm_pois_free_1",
                        feature_name="num_supermarkets"),
          ])

# Export SpellBook into a file
my_spellbook.author = "Juan dela Cruz"
my_spellbook.description = "Good Features for Economic Indicators"
my_spellbook.to_json("my_features.json")
\end{lstlisting}

Once a \texttt{SpellBook} is exported to a file, it can be version-controlled,
shared, and reused to other datasets. In the demonstration below, the
\texttt{Spells} in the exported SpellBook, \texttt{my\_features.json}, will be
casted on a new set of points:

\begin{lstlisting}[language=Python]
from geomancer.spellbook import SpellBook
from test.conftest import sample_points_new

exported_spellbook = SpellBook.read_json("my_features.json")
df = sample_points_new() # load your own data

# Cast someone's Spells into your own data
df_with_features = exported_spellbook.cast(df)
\end{lstlisting}

\section{Case study: property value estimation in Singapore}

\begin{figure}[!ht]
    \begin{center}
        \includegraphics[width=0.85\linewidth]{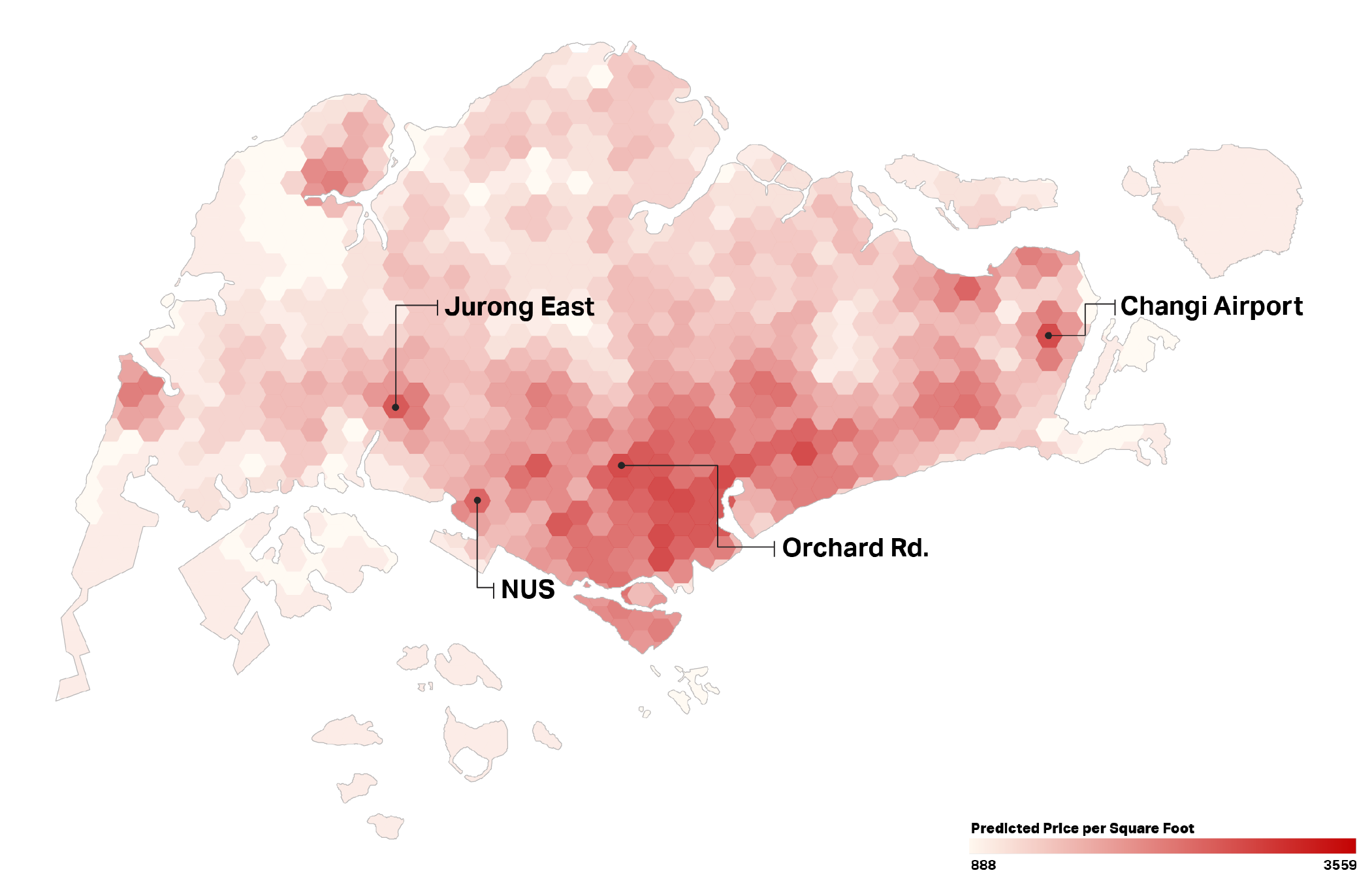}
    \end{center}
    \caption{
        Predictions from a model trained on transaction prices of private
        residential properties based on proximity to certain
        points-of-interest. We treat predicted residential price as an
        \textit{attractiveness index} to quantify and visualize how attractive
        different areas in Singapore are.
    }
  \label{usecase:heatmap}
\end{figure}

We used Geomancer to predict residential prices per square foot in Singapore. 
The raw data was acquired from the Urban Redevelopment Authority's
open listing of apartment and condominium sales in the last
four years \cite{ura2019property}. We used this information to compile a
dataset containing the locations and unit price per square foot for over $50,000$
transactions.

Thus, we are given a raw dataset $\mathcal{D}^{\prime} = \left\{\mathcal{V},
\mathcal{Y}\right\}$, where $\mathcal{V}$ is the property's spatial coordinate
in EPSG:4326, and $\mathcal{Y} \in \mathbb{R}$ is the unit price per square
foot. We then used Geomancer, coupled with OSM data, to define logical features
$\mathcal{F}^{\mathcal{D}}$ such as the number of restaurants within 3-km,
distance to the nearest bus stop, or distance to the nearest nightclub. This
resulted to a feature matrix $X^{\mathcal{D}}$ that will be used for model
training. 

\begin{figure}[h]
    \begin{center}
        \includegraphics[width=0.95\linewidth]{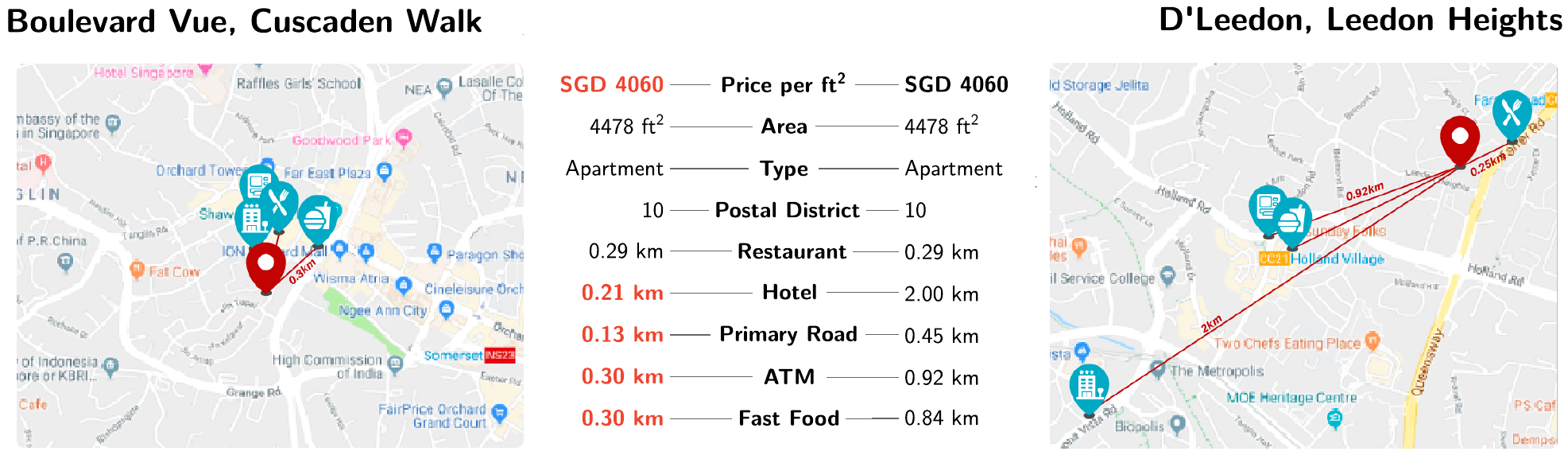}
    \end{center}
    \caption{
        Even if the apartments are of the same size and postal district, the
        more luxurious apartment, Boulevard Vue, is closer to a hotel, primary
        road, ATM and a fast-food restaurant. Accordingly, it is five times
        more expensive than D'Leedon.
    }
  \label{usecase:comparison}
\end{figure}

For this dataset, we were able to extract $388$ geospatial features using
points-of-interests in OpenStreetMap (OSM) and logical features from Geomancer.
After performing a $70-30$ holdout split, we fed these features to a random
forest regressor model \cite{breiman2001random, geurts2006extremely} to predict
the unit price per square foot. Running the trained model on aggregated areas
of Singapore produced a heatmap as shown in Figure \ref{usecase:heatmap}. It is
apparent that the value of an area increases when it's nearer to common
points-of-interests. By zooming-in to two selected properties in Figure
\ref{usecase:comparison}, we can see how the features obtained from Geomancer
directly influenced a property's value.

\begin{figure}[t]
    \begin{subfigure}{0.49\textwidth}
        \centering
        \includegraphics[width=\linewidth]{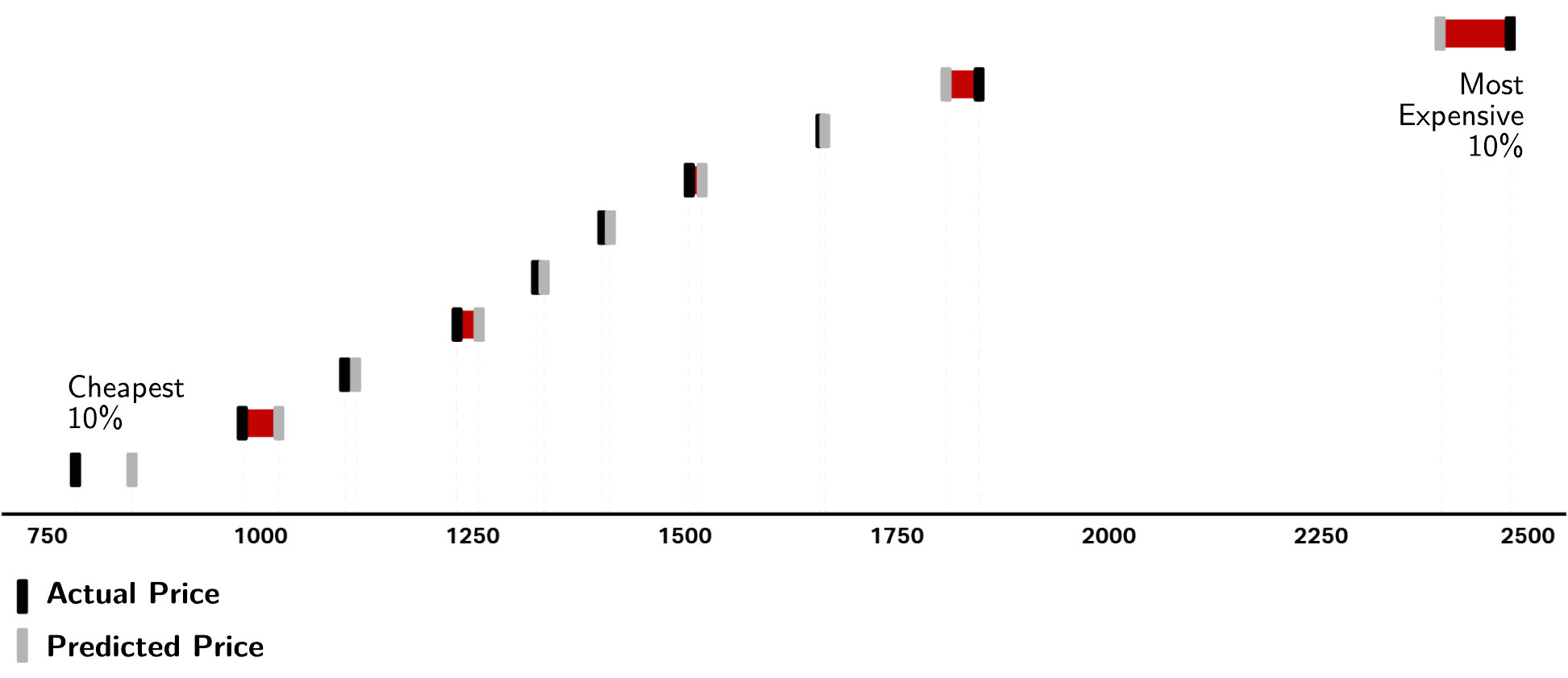}
        \caption{Error-margin}
        \label{usecase:error}
    \end{subfigure}
    \begin{subfigure}{0.49\textwidth}
        \centering
        \includegraphics[width=\linewidth]{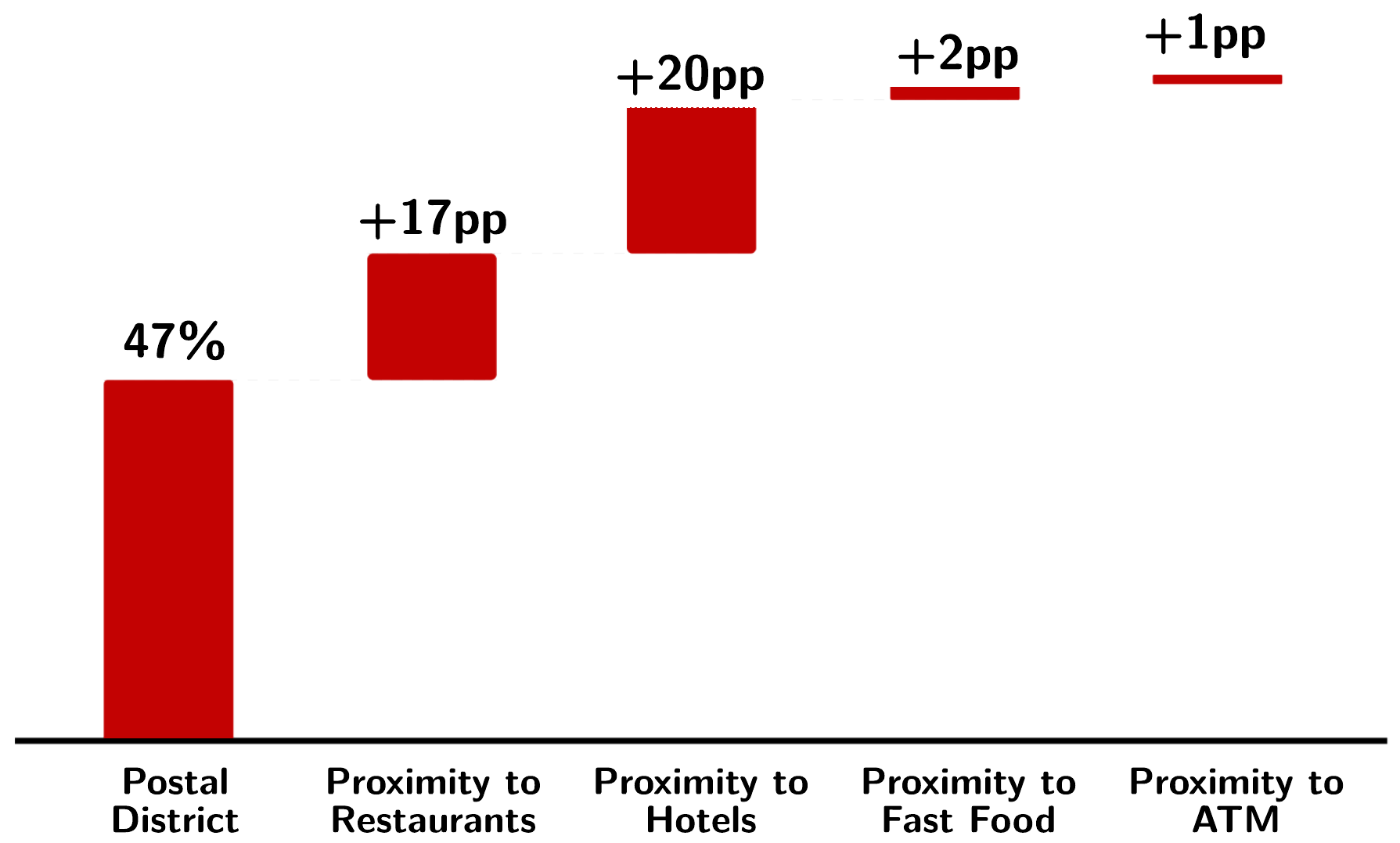}
        \caption{Feature importance}
        \label{usecase:feature}
    \end{subfigure}
    \caption{
        \textit{(Left)} Error margin for the model's predicted price of various
        properties in Singapore. On average, predicted prices are within SGD
        100 of actual prices, even for units on higher or lower end.
        \textit{(Right)} Using different geospatial features allow the model to
        explain up to $87\%$ of real estate value.
    }
\end{figure}

We also compared model predictions to actual selling prices of each property.
Using Geomancer-based features, the model was able to perform really well on
medium prices, but underestimated properties that are extremely cheap or
expensive. Still, we have reliable predictions with error not exceeding SGD $100$
per square foot as shown in Figure \ref{usecase:error}. 

Lastly, we used Geomancer-based features to explain how the model predicts a
property's price. From Figure \ref{usecase:feature}, we can see that over half
of the predictive power comes from the postal district where a unit is located.
Within each district, however, it is possible to improve our price estimation
by $40$ percentage points ($pp$) by adding Geomancer-based features such as
distance to restaurants, hotels, and ATMs. For more information, an interactive
map can be accessed through this
link\footnote{\url{https://thinkdatasci.carto.com/builder/ed486c74-f19e-4a51-862d-a117785e121c}}.

\section{Conclusion}

In this paper, we introduced Geomancer, an open-source framework to perform
geospatial feature engineering at scale. We described the Spell, a logical
feature that serves as the basic building-block of the framework. Then, we
showed how it integrates with the overall architecture and demonstrated how it
can be used through the Python client library. Lastly, we provided a sample
production use-case of Geomancer for predicting residential prices per square
foot in Singapore. Using only Geomancer-based features and OpenStreetMap data,
we were able to achieve $86\%$ accuracy with an error margin of SGD 100. 

For future research, we plan to evaluate user-efficiency and system robustness
in more detail. Finally, we also hope to expand the number of database
connections (e.g. Amazon Athena, Redshift, etc.) and primitive features to
accommodate different cloud providers and other advanced use-cases.

\subsubsection*{Acknowledgments}

This work was supported by the UNICEF Innovation Fund. We would like to thank
our mentors for the insightful discussions and valuable guidance. We would also
like to thank Tiffani Gamboa, Cara Evangelista, and Niek van Veen for the
Singapore case study.
 
\bibliographystyle{unsrt}

\bibliography{bibliography}

\end{document}